\documentclass[10pt,twocolumn,letterpaper]{article}

\usepackage{3dv}
\usepackage{times}
\usepackage{epsfig}
\usepackage{graphicx}
\usepackage{amsmath}
\usepackage{amssymb}

\usepackage{booktabs}
\usepackage{subfig}
\usepackage{graphicx}
\usepackage{url}
\usepackage{arydshln}
\usepackage{xcolor}

\threedvfinalcopy 

\def\threedvPaperID{77} 

\ifthreedvfinal\fi
\begin{document}

\title{Push-the-Boundary: Boundary-aware Feature Propagation for Semantic Segmentation of 3D Point Clouds}

\author{Shenglan Du, Nail Ibrahimli, Jantien Stoter, Julian Kooij, Liangliang Nan\\
Delft University of Technology\\
Julianalaan 134, Delft 2628BL, The Netherlands\\
{\tt\small \{shenglan.du, n.ibrahimli, j.e.Stoter, j.f.p.kooij, liangliang.nan\}@tudelft.nl}
}

\maketitle
\thispagestyle{empty}

\begin{abstract}
Feedforward fully convolutional neural networks currently dominate in semantic segmentation of 3D point clouds.
Despite their great success, they suffer from the loss of local information at low-level layers, posing significant challenges to accurate scene segmentation and precise object boundary delineation.
Prior works either address this issue by post-processing or jointly learn object boundaries to implicitly improve feature encoding of the networks.
These approaches often require additional modules which are difficult to integrate into the original architecture. 

To improve the segmentation near object boundaries, we propose a boundary-aware feature propagation mechanism.
This mechanism is achieved by exploiting a multi-task learning framework that aims to explicitly guide the boundaries to their original locations.
With one shared encoder, our network outputs (i) boundary localization, (ii) prediction of directions pointing to the object's interior, and (iii) semantic segmentation, in three parallel streams.
The predicted boundaries and directions are fused to propagate the learned features to refine the segmentation.
We conduct extensive experiments on the S3DIS and SensatUrban datasets against various baseline methods, demonstrating that our proposed approach yields consistent improvements by reducing boundary errors. Our code is available at https://github.com/shenglandu/PushBoundary.
\end{abstract}

\section{Introduction}
Semantic segmentation of 3D point clouds aims to assign each point a semantic category (\eg, building, tree, window), which is a fundamental yet challenging task in 3D computer vision.
Successful interpretation of semantics from 3D point clouds serves as a crucial prerequisite for various applications, such as autonomous driving~\cite{yurtsever2020survey}, robotics~\cite{crespo2020semantic}, and urban environment modeling~\cite{biljecki2015applications}.

\begin{figure}[t]
	\begin{center}
		\includegraphics[width=0.9\linewidth]{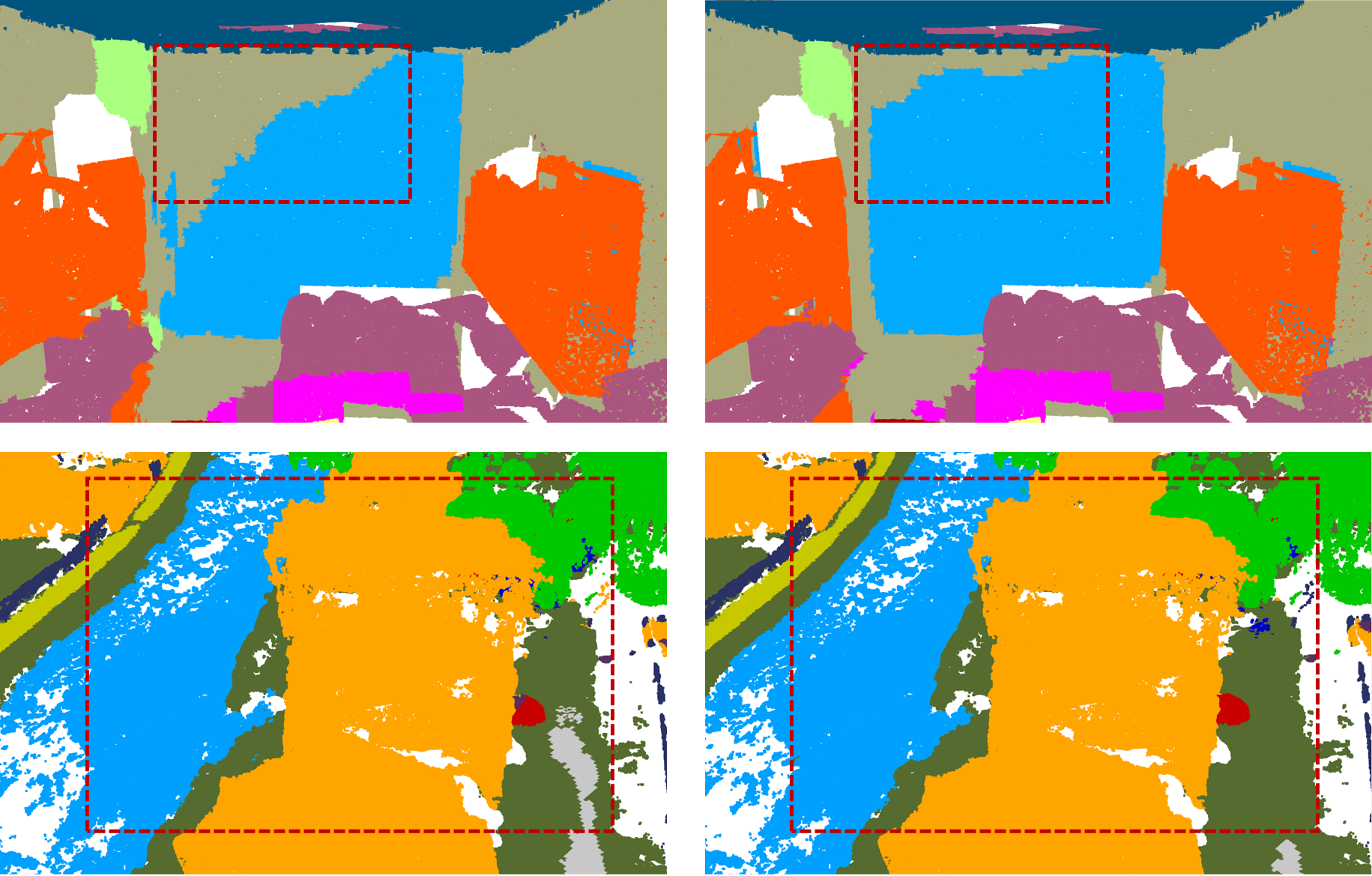}
	\end{center}
	\vspace*{-4.5mm}
	\caption{Comparison between the segmentation results from the previous FCN-based network~\cite{thomas2019kpconv} and ours on both indoor (top) and outdoor (bottom) scenes. Left: standard FCNs lose boundary details at the object level. Right: our approach refined the segmentation at object boundaries.}
	\vspace*{-2mm}
	\label{fig:boundary}
\end{figure}

Driven by the success of deep learning and Fully Convolutional Networks (FCNs) in 2D image recognition, many works have been proposed for semantic segmentation of 3D point clouds by transforming points into regular grids or voxels as input to standard FCNs~\cite{guo2016multi,maturana2015voxnet,riegler2017octnet,su2015multi,wu20153d,yang2019learning}. 
Such methods introduce extra computational costs and information loss which leads to suboptimal performance.
To avoid that, the seminal PointNet~\cite{qi2017pointnet} directly consumes point clouds and extracts point features through a sequence of shared Multi-Layer Perceptrons (MLPs).
Following PointNet, a number of point-based deep learning frameworks have been introduced~\cite{hu2020randla,landrieu2018large,li2018pointcnn,thomas2019kpconv,wang2018deep}.
Despite their strong performances in the semantic segmentation task, they suffer from an often-overlooked limitation in feature propagation: the loss of local information in decoding.
Specifically, since most existing networks adopt FCN-like architectures with the encoding-decoding strategy, the existence of pooling layers in the encoder can capture hierarchical semantic features with increased receptive fields.
Though beneficial for object-level recognition, it leads to coarse feature maps at lower resolutions.
In the decoder, these coarse features are then propagated back to the original resolution via nearest-neighbor upsampling, which ignores point-level variations among different semantic categories.
As a result, networks lose object boundary details and fail to generate accurate predictions.
Figure~\ref{fig:boundary} illustrates the segmentation outputs with blurred object boundaries.

Several works have made fruitful attempts to address semantic segmentation in both 2D and 3D domains. One attempt is to construct contextual affinity through graphical models such as Markov Random Fields (MRFs) and Conditional Random Fields (CRFs)~\cite{chen2017deeplab,zheng2015conditional}.
However, these methods introduce MRFs and CRFs as additional modules that are challenging to be integrated into the networks.
Considering that boundaries play an important role in semantic segmentation, as they naturally indicate the transition between objects of different semantic categories, another line of methods exploits boundary information in the networks~\cite{ding2019boundary,gong2021boundary,hayder2017boundary,hu2020jsenet,zhen2020joint}.
Most of these methods incorporate boundary detection as an auxiliary branch for semantic segmentation.
With one shared encoder, the two tasks implicitly improve each other.
However, such methods do not seek to explicitly tackle semantic segmentation.
Another drawback is that extra encoding layers are needed to merge features from the two tasks, which is more challenging for the network to optimize as more parameters are involved.

Contrary to existing works, our work is mainly motivated by the attempt to \textit{refine semantic segmentation by explicitly pushing the boundary towards desired directions}.
Predicting a direction scheme to refine semantic segmentation has been recently studied in 2D image recognition~\cite{mazzini2019spatial,yuan2020segfix}, although in these works directions mainly serve as post-processing tools to refine generated semantic labels.
To overcome such limitations, we propose a novel end-to-end framework for joint boundary detection, direction prediction, and semantic segmentation.
The proposed network has one FCN feature encoder while jointly giving three streams of point-wise predictions:
(i) a boundary label (i.e., binary prediction),
(ii) a direction vector that originates from the closest boundary and points to the object's interior,
and (iii) a semantic class label.
We demonstrate that even though the FCN architecture is primarily optimized for semantic segmentation, it does provide discriminative features for boundary detection and direction prediction.

The key to our framework is a lightweight guiding mechanism that effectively fuses the boundary and direction priors to refine the segmentation.
Our motivation is to reduce the loss of local information in the decoder of the framework.
We propose to guide the feature propagation using the predicted direction (i.e., pointing from boundary to interior) near object boundaries.
In such a way, feature mixture from different semantic classes is prevented while semantic boundaries are explicitly pushed along the desired direction. Our contributions can be summarized in two folds:

\begin{itemize}
	\setlength\itemsep{0.05em}
	\item To the best of our knowledge, the proposed network is the first end-to-end framework for joint semantic segmentation, boundary detection, and direction prediction in the 3D domain.
	The tasks of boundary detection and direction prediction can appropriately improve segmentation output.
	\item We introduce a novel boundary-aware feature upsampling strategy to guide feature propagation towards the predicted directions near object boundaries, which can be easily plugged into existing frameworks.
\end{itemize}

\section{Related Work}
\subsection{Point Cloud Semantic Segmentation}
Several methods have been proposed for semantic segmentation of point clouds.
They can be roughly divided into projection-based, voxel-based, and point-based methods.

\textit{Projection-based methods} first project points onto specified 2D planes and then apply 2D FCNs to recognize 3D objects~\cite{guo2016multi,su2015multi,yang2019learning,yu2018multi}.
These methods do not fully utilize the data due to the loss of information in the $z^{th}$ dimension in 2D projection, limiting their ability to analyze complex 3D objects in 3D scenes.

\textit{Voxel-based methods} discretize point clouds over a volumetric 3D grid for shape classification and semantic segmentation using standard 3D FCNs~\cite{maturana2015voxnet,riegler2017octnet,wu20153d}.
Such methods require data transformation, which introduces extra computational costs and information loss due to the limited resolution of the grid structure.
Besides, the voxel representation also has a high demand for GPU memory, limiting its application to single objects and small scenes.

Compared with the above two types of methods, \textit{point-based methods} directly take point clouds as input to the networks and have demonstrated promising performances on various datasets.
PointNet~\cite{qi2017pointnet} is the first successful attempt in this direction, which adopts a sequence of MLPs and max-pooling operators to learn global features that can be used for 3D shape classification and scene segmentation.
PoinNet++~\cite{qi2017pointnet++} recursively applies PointNet over a nested partitioning of the points to learn fine-grained details of local geometrical structures.
Besides MLPs, a number of works~\cite{hua2018pointwise,li2018pointcnn,thomas2019kpconv,wang2018deep} explore 3D convolutional operators on point clouds.
Another line of work has also exploited Graph Convolutional Networks (GCNs) and self-attention mechanisms for gaining contextual knowledge of the points from local neighborhood graph patches~\cite{jiang2019hierarchical,landrieu2018large,simonovsky2017dynamic,wang2019graph,zhao2021point}.
Our work is built upon FCN-like architectures such as PointNet++~\cite{qi2017pointnet++} and KP-Conv~\cite{thomas2019kpconv}, aiming to tackle the loss of local boundary details in these networks.

\begin{figure*}
	\begin{center}
		\includegraphics[width=0.92\linewidth]{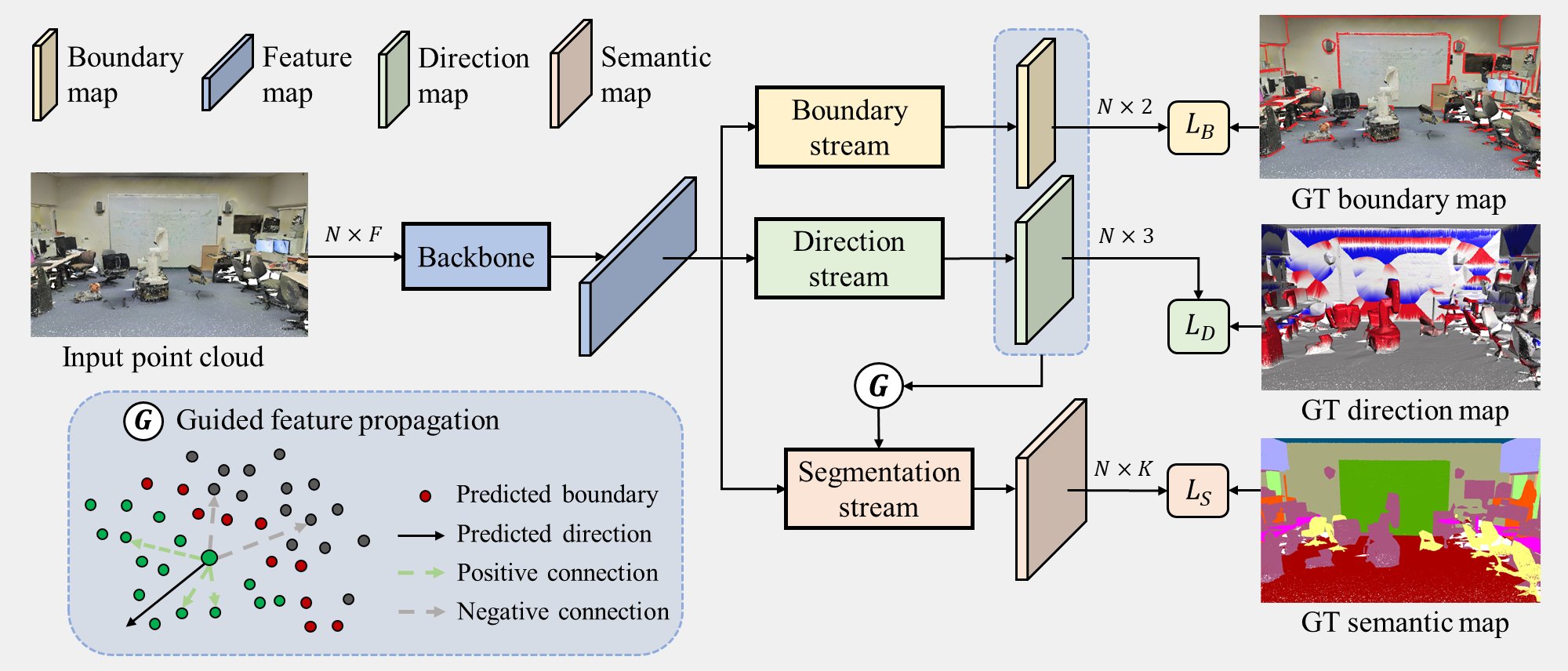}
	\end{center}
	\vspace*{-4.5mm}
	\caption{An overview of the network architecture. Our network can adopt various FCN-like feature encoders (\eg, KP-Conv~\cite{thomas2019kpconv}, PointNet++~\cite{qi2017pointnet++}) for learning point features.
	$N$ denotes the number of input points, $F$ the input feature dimension, $K$ the output category numbers, and GT the ground truth. $L_B$, $L_D$, and $L_S$ are the network losses for the boundary, direction, and segmentation streams, respectively.
	In the guided feature propagation, a positive connection means the vector of the point pair has a positive cosine angle with the predicted direction, while a negative connection means a negative cosine angle.}
	\vspace*{-2mm}
	\label{fig:overview}
\end{figure*}

\subsection{Boundary-aware Segmentation}
Standard FCN-like networks for semantic segmentation fail to model point-level accurate object boundaries, mainly due to the significant loss of low-level features both in the feature encoding and decoding layers.
Many works have been introduced to sharpen the object boundaries.

In 2D image recognition, Boundary Neural Field~\cite{bertasius2016semantic} formalizes a global energy model to enhance semantic segment coherence with predicted boundary cues.
Other works combine the semantic segmentation task and the boundary detection task into one network~\cite{ding2019boundary,hayder2017boundary,liu2017richer,su2019selectivity,takikawa2019gated,yu2017casenet}, in which the two tasks share a common feature encoder and are expected to mutually improve each other.

In the 3D domain, boundary-aware strategies have also been adopted by a few works.
JSENet~\cite{zhen2020joint} employs extra feature enhancement modules that require a curriculum learning strategy for good segmentation performance.
CBL~\cite{tang2022contrastive} introduces additional loss terms to contrast the features across the scene boundaries.
Contrary to them, our work focuses on interior information passage in network decoding.
We show that the guided feature propagation naturally recovers information near boundaries, which does not require exterior constraints or additional training modules.
Several works~\cite{gong2021boundary,xu2021investigate} address 3D boundary detection by adaptive methods or pre-process modules.
Instead, our method does not treat boundaries as intermediate results. It jointly learns boundary cues for guiding the feature propagation, which can be easily plugged into existing networks.

\subsection{Graphical Models for Segmentation}
In the 2D domain, some works have proposed to integrate graphical models, such as MRFs and CRFs, into the networks for segmentation refinement~\cite{chen2017deeplab,krahenbuhl2011efficient,zheng2015conditional}.
Based on the fact that pixels presenting similar features tend to have the same semantic labels, this line of work formulates segmentation as a probabilistic inference problem using graphical models.
While these methods can sharpen the segmentation mask near object boundaries, integrating additional MRFs and CRFs into the networks is still challenging.
Besides, this strategy requires extra computational costs and is therefore slow in speed.

\section{Methodology}
Our network consists of one FCN feature encoder followed by branches of three task streams:
(i) boundary detection,
(ii) direction prediction (i.e., for each point, we predict a direction vector from its closest boundary to the object's interior),
and (iii) semantic segmentation.
The outputs of (i) and (ii) are then fused to guide the feature propagation in a boundary-aware manner. Figure~\ref{fig:overview} overviews our architecture. In the following, we explain each part in detail.

\subsection{Architecture}
We use KP-Conv~\cite{thomas2019kpconv} as our backbone network, which directly applies fully convolutional layers over 3D point clouds using kernel-point convolution.
In Section~\ref{subsec:s3dis}, we also provide segmentation experiments adopting other baseline networks as backbones, such as PointNet++~\cite{qi2017pointnet++}.

\textbf{Boundary detection stream}.
For the input point cloud $P\in R^{N\times F}$, where $N$ is the number of points and $F$ is the input feature dimension, this stream predicts a binary map $B\in R^{N\times 2}$, with 1 for the points on boundaries and 0 for interior points.
Note that we refer to semantic boundaries (i.e., boundaries between different semantic categories) in this task.
Due to that boundary points only account for a small portion of the whole point set, we use the weighted binary cross-entropy loss to supervise this task.

\textbf{Direction prediction stream}.
Predicting a direction scheme to refine semantic segmentation has been recently studied~\cite{mazzini2019spatial,yuan2020segfix} in the 2D domain.
These works require separate learning for direction vectors, which then serve as a post-processing technique to refine the segmentation output.
In this work, we extend the direction scheme to the 3D domain and learn it in an end-to-end manner.
The direction stream predicts a direction map $D\in R^{N\times 3}$, with each row $(d_x, d_y, d_z)$ giving a unit direction vector pointing from the closest boundary to the object's interior.
Designed in this way, the learned $D$ generates for each point a pointer towards its homogeneous region interior.
We consider direction prediction as a regression task and thus adopt Mean Squared Error loss for supervision.
The predicted directions are further fused with boundary predictions to guide the feature propagation in the semantic segmentation task stream.

\textbf{Semantic segmentation stream}.
Given a point cloud $P\in R^{N\times F}$, the semantic segmentation branch outputs a semantic mask $S\in R^{N\times K}$, where $K$ is the total number of semantic categories.
This stream predicts a probability distribution for each point over $K$ categories. We supervise this task using the standard cross-entropy loss.

\begin{figure*}
	\begin{center}
		\includegraphics[width=0.95\linewidth]{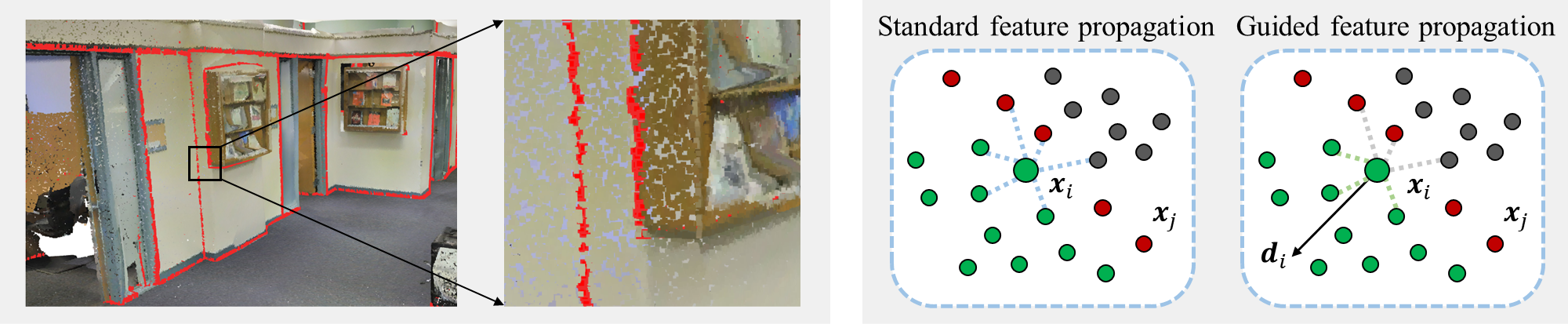}%
	\end{center}
	\vspace*{-4.5mm}
	\caption{Feature propagation near object boundaries. Left: drastic semantic transitions happen across the object boundaries within a small local neighborhood. Right: two ways of feature propagation. We propagate features from neighboring points $\mathbf{x_j}$ in the higher layer to the center $\mathbf{x_i}$ in the lower layer. \textit{Standard feature propagation} ignores object boundaries (red dots), which leads to the mixed feature representation of $\mathbf{x_i}$. In contrast, \textit{our proposed strategy} uses the predicted direction vector $\mathbf{d_i}$ to guide the feature propagation. Features from neighboring $\mathbf{x_j}$ aligned with $\mathbf{d_i}$ are encouraged to propagate to the center $\mathbf{x_i}$ (green dashed lines), while feature propagation from the opposed neighbors is constrained (grey dashed lines).}
	\vspace*{-2mm}
	\label{fig:guidefeat}
\end{figure*}

\subsection{Boundary-aware Feature Propagation}
\label{subsec:guidefeat}
FCNs adopt pooling layers that progressively sub-sample points and obtain high-level semantic features in the latent space.
In the subsequent decoding stage, features are propagated from sub-sampled points to the original points.
A common propagation strategy is based on inverse distance interpolation in a local $k$-Nearest Neighborhood (kNN).
For example, PointNet++~\cite{qi2017pointnet++} uses $k=3$ for feature propagation, while KP-Conv~\cite{thomas2019kpconv} uses $k=1$ that reduces to the nearest neighbor feature upsampling.
Standard feature propagation assumes that spatially close points also share the semantic affinity.
However, it ignores drastic semantic transitions across object boundaries.
Points at the boundary can inherit information from different objects, thus resulting in blurred feature representation, which is not favorable for the final segmentation.

We use the predicted boundaries and directions to guide the feature propagation in decoding layers, where features are encouraged to pass along the desired direction for generating purified feature maps.
The guided feature propagation is illustrated in Figure~\ref{fig:guidefeat}.
In the FCN decoder, features are up-sampled from sparser points in deep layers to denser points in shallow layers.
We propagate features of points $\mathbf{x_j}$ in $l^{th}$ layer to points $\mathbf{x_i}$ in $l-1^{th}$ layer. The feature propagation is performed by
\begin{equation} \label{eq:1}
	f^{l-1}(\mathbf{x_i})=\sigma (\Phi(\frac{\sum_{j=1}^{k}w(\mathbf{x_j})f^{l}(\mathbf{x_j})}{\sum_{j=1}^{k}w(\mathbf{x_j})})\oplus f_{skip}^{l-1}(\mathbf{x_i})),
\end{equation}
where $\mathbf{x}\in R^3$ denotes a 3D point and $k$ is the total number of neighboring points.
The decoder features $f^{l}(\mathbf{x_j})$ of neighboring points $\mathbf{x_j}$ at $l^{th}$ layer are first interpolated based on corresponding weights $w$, and then concatenated with skip-linked encoder features that will pass through a MLP layer and a Relu activation function to obtain the new features $f^{l-1}(\mathbf{x_i})$ for $\mathbf{x_i}$.
$f_{skip}$ denotes the skip-linked features, $\Phi(\cdot)$ a MLP operator, and $\sigma(\cdot)$ the ReLU activation function.
We compute an adaptive weight term $w$ to guide the feature propagation process,
\begin{flalign} \label{eq:2}
	\quad w(\mathbf{x_j}) =& \max(0, w_{s}(\mathbf{x_j},\mathbf{x_i}) + \alpha w_{c}(\mathbf{x_j},\mathbf{x_i})), \\
	w_{s}(\mathbf{x_j} ,\mathbf{x_i}) & = \exp{\frac{-\vert\vert \mathbf{x_j} - \mathbf{x_i}\vert\vert_2}{r}} \nonumber, \\
	w_{c}(\mathbf{x_j},\mathbf{x_i}) & = \exp(P_b(\mathbf{x_i})-1)\nonumber \cos(\mathbf{x_j}-\mathbf{x_i},\mathbf{d_i}),
\end{flalign}
where we linearly combine the spatial similarity $w_s$ and cosine similarity $w_c$ between $\mathbf{x_j}$ and $\mathbf{x_i}$.
$P_b (\mathbf{x_i})$ is the predicted boundary probability of point $\mathbf{x_i}$, $\mathbf{d_i}$ the predicted direction of $\mathbf{x_i}$, and $\cos(\cdot)$ the cosine angle of two vectors.
$\max(\cdot)$ is used for eliminating negative weights generated in the computation. The constant coefficients $r$ and $\alpha$ are used to balance the two terms.
The computed weights are further normalized, i.e.,
\begin{equation} \label{eq:3}
	w(\mathbf{x_j})=\frac{w(\mathbf{x_j})}{\sum_{j=1}^{k}{w(\mathbf{x_j})}}.
\end{equation}

Our propagation strategy favors semantic segmentation in three folds.
(i) For points near boundaries (i.e., with $P_b\approx 1$), we assign significant importance to the neighbors aligned with the direction while eliminating the influence of opposed neighbors, encouraging the features to propagate along the desired direction.
(ii) For interior points (i.e., with $P_b\approx 0$), the second cosine term is largely reduced.
All neighbors majorly contribute to feature propagation according to their distances to the center. Therefore, we facilitate smooth segmentation in the object's interior. 
(iii) Our designed weight term is a continuous function over $B$ and $D$, which helps the gradient backpropagation of the network.
In Section~\ref{subsec:ablation}, we conduct ablation studies to further verify the effectiveness of the proposed boundary-aware feature propagation mechanism.

\subsection{Network Supervision}
Our network involves three tasks (see Figure~\ref{fig:overview}), which are supervised using the three loss terms, respectively.

\textbf{Boundary detection loss}.
Boundary detection is a binary classification task. Due to the sparsity of boundary points in the dataset, we adopt the weighted binary cross entropy loss to supervise this task, 
\begin{equation} \label{eq:4}
	L_B=-\sum_{i=1}^N [\,\beta \hat{b_i} \log b_i+(1-\beta)(1-\hat{b_i})\log(1-b_i)],
\end{equation}
where $\hat{b_i}$ denotes the GT binary label, i.e., 1 for boundary points and 0 for interior points. $b_i$ is the network softmax output of the $i^{th}$ point.
We use a coefficient $\beta$ to balance the boundary class and the object's interior class.

\textbf{Direction prediction loss}.
Unlike previous work of 2D image segmentation~\cite{yuan2020segfix} that predicts discrete directions by evenly dividing the entire direction into a set of ranges, we perform continuous prediction in the whole space.
Therefore, our predictions are more adaptable for complex 3D scenes.
Specifically, we tackle direction prediction as a regression task and adopt MSE loss for supervision, i.e.,
\begin{equation} \label{eq:5}
	L_D=\sum_{i=1}^N \| \mathbf{d_i}-\hat{\mathbf{d_i}} \|_2^2,
\end{equation}
where $\mathbf{d_i}\in R^3$ is the predicted direction, and $\hat{\mathbf{d_i}}\in R^3$ is the GT direction (both directions have a unit magnitude).
We also tested with the dot product loss of $\mathbf{d_i}$ and $\hat{\mathbf{d_i}}$, while our experiments suggested that the MSE loss gives robust direction predictions that are less variant to dataset scales.

\textbf{Semantic segmentation loss}.
We use the standard cross entropy loss to supervise the semantic segmentation stream,
\begin{equation} \label{eq:6}
	L_S=-\sum_{i=1}^N y_i^s \log p^s (\mathbf{x_i}),
\end{equation}
where $y_i^s \in R^K$ denotes the one-hot vector of the Ground Truth (GT) semantic label $s$ of the $i^{th}$ point. $p^s (\mathbf{x_i})$ is the predicted probability of the $i^{th}$ point over the GT category obtained from the network softmax layer.

Accordingly, the total network loss is given by
\begin{equation} \label{eq:7}
	L=L_S+\lambda_1 L_B+\lambda_2 L_D,
\end{equation}
where we use $\lambda_1$ and $\lambda_2$ to balance between different losses.

\subsection{Implementation Details and Parameters}
Our method is implemented in Pytorch~\cite{paszke2019pytorch}.
Regarding the hyperparameters, we set $\alpha = 1.0$ and $r=0.125$ for the guided feature propagation in Equation~\ref{eq:2}.
For supervising the boundary stream in Equation~\ref{eq:4}, we set $\beta=0.6$. For the weights in Equation~\ref{eq:7}, we set $\lambda_1=3.0$ and $\lambda_2=0.3$.

\section{Experiments}

\begin{table*}
	\begin{center}
		\scalebox{0.75}{
			\renewcommand{\arraystretch}{1.2}
			\begin{tabular}{r|cc|ccccccccccccc}
				\hline
				Method & OA(\%) & \textbf{mIoU(\%)} & ceiling & floor & wall & beam & col. & window & door & table & chair & sofa & book. & board & clutter \\
				\hline\hline
				PointNet++~\cite{yan2019pointnetpytorch} (w/o N) & 83.5 & 53.6 & 89.6 & \textbf{97.6} & 74.6 & 0.0 & 4.7 & 55.1 & \textbf{19.9} & 78.2 & 68.9 & \textbf{66.3} & 41.7 & 55.6 & \textbf{44.2} \\
				+ Ours & \textbf{83.6} & \textbf{53.9} & \textbf{89.7} & 97.2 & \textbf{75.4} & 0.0 & \textbf{7.4} & \textbf{60.8} & 14.1 & \textbf{79.0} & \textbf{69.5} & 65.8 & \textbf{42.1} & \textbf{57.0} & 42.6 \\
				\hdashline
				PointNet++~\cite{yan2019pointnetpytorch}  (w/ N) & 83.9 & 53.9 & 91.4 & 96.6 & 76.0 & 0.0 & 8.0 & \textbf{53.7} & 16.9 & 81.8 & \textbf{70.5} & 63.8 & 48.9 & 49.4 & 43.6 \\
				+ Ours & \textbf{84.3} & \textbf{55.1} & \textbf{91.5} & \textbf{97.2} & 76.0 & 0.0 & \textbf{13.8} & 53.4 & \textbf{19.1} & \textbf{83.6} & 70.4 & \textbf{65.4} & \textbf{49.6} & \textbf{50.3} & \textbf{45.4} \\	
				\hline
				KP-Conv rigid~\cite{thomas2019kpconv} (w/o N) & - & 65.4 & 92.6 & 97.3 & 81.4 & 0.0 & 16.5 & 54.5 & 69.5 & \textbf{80.2} & 90.1 & 66.4 & \textbf{74.6} & \textbf{63.7} & 58.1 \\
				+ Ours & 89.7 & \textbf{67.1} & \textbf{94.0} & \textbf{97.9} & \textbf{82.6} & 0.0 & \textbf{23.3} & \textbf{56.6} & \textbf{75.4} & 80.1 & \textbf{91.1} & \textbf{75.7} & 74.4 & 62.3 & \textbf{59.1} \\
				\hdashline
				KP-Conv rigid~\cite{thomas2019kpconv}  (w/ N) & 89.2 & 65.6 & \textbf{94.0} & 97.9 & 81.6 & 0.0 & 20.0 & 54.5 & 64.4 & 80.1 & 91.6 & 77.4 & 73.8 & 59.2 & 59.0 \\
				+ Ours & \textbf{89.6} & \textbf{67.2} & 93.9 & 97.9 & \textbf{82.7} & \textbf{0.2} & \textbf{23.8} & \textbf{55.0} & \textbf{73.7} & \textbf{80.5} & \textbf{91.8} & \textbf{77.7} & \textbf{74.1} & \textbf{63.2} & 59.0 \\
				\hline			
		\end{tabular}}
	\end{center}
	\vspace*{-4.5mm}
	\caption{
	Semantic segmentation results achieved on the S3DIS dataset~\cite{armeni20163d}, following instructions of the officially released code of each method.
	These methods are trained using either 3D coordinates and color information (w/o N) or with additional normal information (w/ N).
	We adopt unweighted cross-entropy loss for all the segmentation experiments to achieve a fair comparison.
	Overall Accuracy (OA, \%), mean Intersection over Union (mIoU, \%), and per-category IoU scores are reported.
	We report averaged scores over five training runs for PointNet++~\cite{yan2019pointnetpytorch} based networks due to performance variations.
	}
	\vspace*{-2mm}
	\label{tab:s3dis}
\end{table*}

\textbf{Evaluation setup}.
We evaluate the effectiveness of our approach in semantic segmentation of 3D point clouds using two state-of-the-art network architectures as our backbone: PointNet++~\cite{qi2017pointnet++} and KP-Conv~\cite{thomas2019kpconv}.
Pointnet++ recursively applies point-wise MLP operators over a nested partitioning of the point clouds to obtain hierarchical point features.
In contrast, KP-Conv explores point convolution in 3D Euclidean space.
Both approaches employ FCN-like architectures with the encoding-decoding strategy.

Our experiments are carried out on datasets of both indoor and outdoor scenes.
We use standard metrics including Overall Accuracy (OA), mean Intersection over Union (mIoU), and per-category IoU scores for the evaluation.
To achieve a fair per-category comparison, we use the standard unweighted cross-entropy loss for all the experiments, and we adopt the same experimental settings as in the original corresponding baseline approaches.

\textbf{Ground-truth boundary maps and direction maps}.
Both GT boundary maps and direction maps are directly derived from the raw dataset. To generate GT boundary maps, we use kNN search for each point in the dataset (we empirically set $k=4$).
If its semantic label differs from any of its neighbors, we recognize it as a boundary point.
To generate GT direction maps, we search the closest boundary point for each point, and the direction is assigned as the direction pointing from the closest boundary point to the current point.
We use the normalized direction vectors (i.e., with a length of 1) as our GT for training.

\subsection{Semantic Segmentation on Indoor Scenes}
\label{subsec:s3dis}
For indoor scenes, we use the challenging Stanford Large-Scale 3D Indoor Spaces (S3DIS) dataset~\cite{armeni20163d} for evaluation.
S3DIS is a large-scale dataset for semantic indoor scene parsing. 
Each point is annotated with a semantic label from 13 categories (e.g., ceiling, table, window).
Since the S3DIS dataset contains massive points that are impractical to be directly segmented, both the two baseline methods were intentionally designed to train and test on downsampled data. We follow the same data sampling strategies as the baseline networks.
Furthermore, to align with the previous works~\cite{landrieu2018large,qi2017pointnet,thomas2019kpconv,zhao2021point}, we also use Area 5 for testing and the rest areas for training.

In Table~\ref{tab:s3dis}, we report the performance comparison between the two baselines and our corresponding networks.
These methods are either trained using only 3D coordinates and color information (w/o N), or using 3D coordinates, color information, and normals (w/ N).
We consider training the networks using normal information as they help discriminate between boundary and non-boundary points.
Compared to the baselines, our boundary-aware feature propagation mechanism achieves consistent improvements in terms of both OA and mIoU scores.
Trained using only 3D coordinates and color information, our method obtains mIoU gains of $0.3\%$ and $1.7\%$ with the backbones PointNet++ and KP-Conv, respectively.
When trained using 3D coordinates, color information, and normals, our method obtains mIoU gains of $1.2\%$ and $1.6\%$ with the backbones PointNet++ and KP-Conv, respectively.
Specifically, we observe significant improvements in categories such as column, window, door, and board, for which accurate boundary delineation is difficult to achieve in the baseline methods.
In some scenarios our method can also propagate segmentation errors to a larger extent, leading to a decrease in mIoU scores of certain categories (e.g., board).
Nevertheless, the qualitative results show that our segmentation outputs more regularized shapes (Section~\ref{subsec:visual}).
In \textit{Supplementary Material}, we present more comparisons with other state-of-the-art works in the S3DIS benchmark.

\subsection{Semantic Segmentation on Outdoor Scenes}
\label{subsec:sensat}

\begin{table*}
	\begin{center}
		\scalebox{0.77}{
			\renewcommand{\arraystretch}{1.2}
			\begin{tabular}{l|cc|ccccccccccccc}
				\hline
				Method & OA(\%) & \textbf{mIoU(\%)} & ground & veg. & building & wall & bridge & parking & rail & traffic. & street. & car & footpath & bike & water \\
				\hline\hline
				PointNet~\cite{qi2017pointnet} & 80.8 & 23.7 & 68.0 & 89.5 & 80.1 & 0.0 & 0.0 & 4.0 & 0.0 & 32.0 & 0.0 & 35.1 & 0.0 & 0.0 & 0.0 \\
				PointNet++~\cite{qi2017pointnet++} & 84.3 & 32.9 & 72.5 & 94.2 & 84.8 & 2.7 & 2.1 & 25.8 & 0.0 & 31.5 & 11.4 & 38.8 & 7.1 & 0.0 & 56.9 \\
				TangentConv~\cite{tatarchenko2018tangent}& 77.0 & 33.3 & 71.5 & 91.4 & 75.9 & 35.2 & 0.0 & 45.3 & 0.0 & 26.7 & 19.2 & 67.6 & 0.0 & 0.0 & 0.0 \\
				SPGraph~\cite{landrieu2018large} & 85.3 & 37.3 & 69.9 & 94.6 & 88.9 & 32.8 & 12.6 & 15.8 & \textbf{15.5} & 30.6 & 23.0 & 56.4 & 0.5 & 0.0 & 44.2 \\
				SparseConv~\cite{graham20183d} & 88.7 & 42.7 & 74.1 & 97.9 & 94.2 & 63.3 & 7.5 & 24.2 & 0.0 & 30.1 & 34.0 & 74.4 & 0.0 & 0.0 & 54.8 \\ 
				RandlaNet~\cite{hu2020randla} & 89.8 & 52.7 & 80.1 & 98.1 & 91.6 & 49.0 & 40.8 & 51.6 & 0.0 & 56.7 & 33.2 & 80.1 & 32.6 & 0.0 & 71.3 \\
				KP-Conv~\cite{thomas2019kpconv} & 93.2 & 57.6 & \textbf{87.1} & \textbf{98.9} & 95.3 & 74.4 & 28.7 & 41.4 & 0.0 & 56.0 & 54.4 & 85.7 & 40.4 & 0.0 & \textbf{86.3} \\
				\hline
				Ours & \textbf{93.8} & \textbf{59.7} & 85.8 & 98.9 & \textbf{96.8} & \textbf{79.3} & \textbf{49.7} & \textbf{52.4} & 0.0 & \textbf{62.1} & \textbf{57.5} & \textbf{86.8} & \textbf{42.0} & 0.0 & 65.5 \\
				\hline			
		\end{tabular}}
	\end{center}
	\vspace*{-4.5mm}
	\caption{Semantic segmentation results achieved on SensatUban dataset~\cite{hu2021towards}, evaluated on the Birmingham block 2, 8,  and the Cambridge block 15, 16, 22, and 27.
	Overall Accuracy (OA, \%), mean Intersection over Union (mIoU, \%), and per-category IoU scores are reported.
	The results of the seven competing networks are from the original SensatUrban benchmark paper~\cite{hu2021towards}.
	To achieve a fair comparison, 
	we train our method using only the 3D coordinates and color information.
	}
	\vspace*{-2mm}
	\label{tab:sensat}
\end{table*}

For outdoor scenes, we use the most recent SensatUrban dataset~\cite{hu2021towards} for evaluation.
SensatUrban is a large-scale UAV photogrammetry point cloud dataset, 
consisting of nearly three billion points with fine-grained semantic labels (\eg, building, high vegetation, rail) collected from three UK cities.
Following the SensatUrban benchmark, we use the Birmingham block 1 and 5, and the Cambridge block 7 and 10 for validation. We use the Birmingham block 2 and 8, and the Cambridge block 15, 16, 22, and 27 for testing.
The rest tiles are used for training. We experiment on SensatUrban using KP-Conv~\cite{thomas2019kpconv} as our backbone.

In Table~\ref{tab:sensat}, we present the performance of our approach in comparison with KP-Conv and other state-of-the-art methods.
Our method has achieved an OA of $93.9\%$ and an mIoU of $59.7\%$, which outperforms all the seminal competitors on this benchmark.
The results demonstrate the capability of our approach to generalize to large-scale outdoor urban scenes.
Compared to the state-of-the-art baseline KP-Conv~\cite{thomas2019kpconv}, using our boundary-aware feature propagation strategy improves its mIoU by a margin of $2.1\%$.
Regarding per category performance, improvements in most categories can also be observed, especially in the categories that exhibit distinct geometric boundaries, such as bridges, roads, and footpaths.
The minority categories such as rail and bike cannot be successfully recognized by most methods listed in the table, which is due to the class imbalance in the SensatUrban dataset.
We also observe a performance drop in the water category ($20.8\%$ IoU compared to baseline).
This is attributed to the fact that water in urban scenes often has irregular geometrical boundaries, which leads to noisy estimates in both boundary detection and direction prediction, yielding less accurate segmentation.

\subsection{Ablation Study}
\label{subsec:ablation}
In this section, we present ablation studies to support our contributions in detail.
We use the S3DIS dataset as we can access the GT semantic labels of all the points. We conduct our ablation studies adopting KP-Conv~\cite{thomas2019kpconv} as our backbone network.
Same as in the previous works~\cite{landrieu2018large,qi2017pointnet,thomas2019kpconv,zhao2021point}, we use Area 5 for testing and the rest areas for training.
All the experiments are conducted under the same hyperparameter settings as in Section~\ref{subsec:s3dis}.

Table~\ref{tab:ablation} presents the results of the ablation experiments.
From this table, we observe that adding the boundary stream to the baseline improves mIoU by $1.0\%$ and OA by $0.1\%$.
Adding the direction stream to the baseline improves mIoU by $0.8\%$ but slightly decreases the OA ($0.1\%$), and combining the three streams with the standard feature propagation achieves a $0.2\%$ gain in OA and $0.7\%$ gain in mIoU over the baseline.
Compared to adding only the boundary stream, adding both the boundary stream and the direction stream to the network shows a decrease of $0.3\%$ in mIoU. Even though, the direction stream brings important information for segmentation refinement in the next step.
Using the proposed boundary-aware feature propagation mechanism, we observe an increase in both metrics, i.e., $0.4\%$ in OA and $1.6\%$ in mIoU.
This reveals that more effective features are inherited to produce purified feature maps. 

\begin{table}
	\begin{center}
		\scalebox{0.85}{
			\renewcommand{\arraystretch}{1.2}
			\begin{tabular}{l|cc}
				\hline
				Method & OA(\%) & \textbf{mIoU(\%)} \\
				\hline\hline
				(1) Baseline network & 89.2 & 65.6 \\
				(2) Baseline + Boundary Stream  & 89.3 & 66.6 \\
				(3) Baseline + Direction Stream & 89.1 & 66.2 \\
				(4) Full network with SFP & 89.4 & 66.3 \\
				(5) Full network with GFP & 89.6 & 67.2 \\
				\hline
		\end{tabular}}
	\end{center}
	\vspace*{-4.5mm}
	\caption{mIoU scores of the ablated networks from (1) to (5). SFP denotes standard feature propagation and GFP denotes guided feature propagation.
	All networks are trained using 3D coordinates, color information, and normals.
	}
	\vspace*{-2mm}
	\label{tab:ablation}
\end{table}

\begin{figure*}
	\begin{center}
		\includegraphics[width=0.95\linewidth]{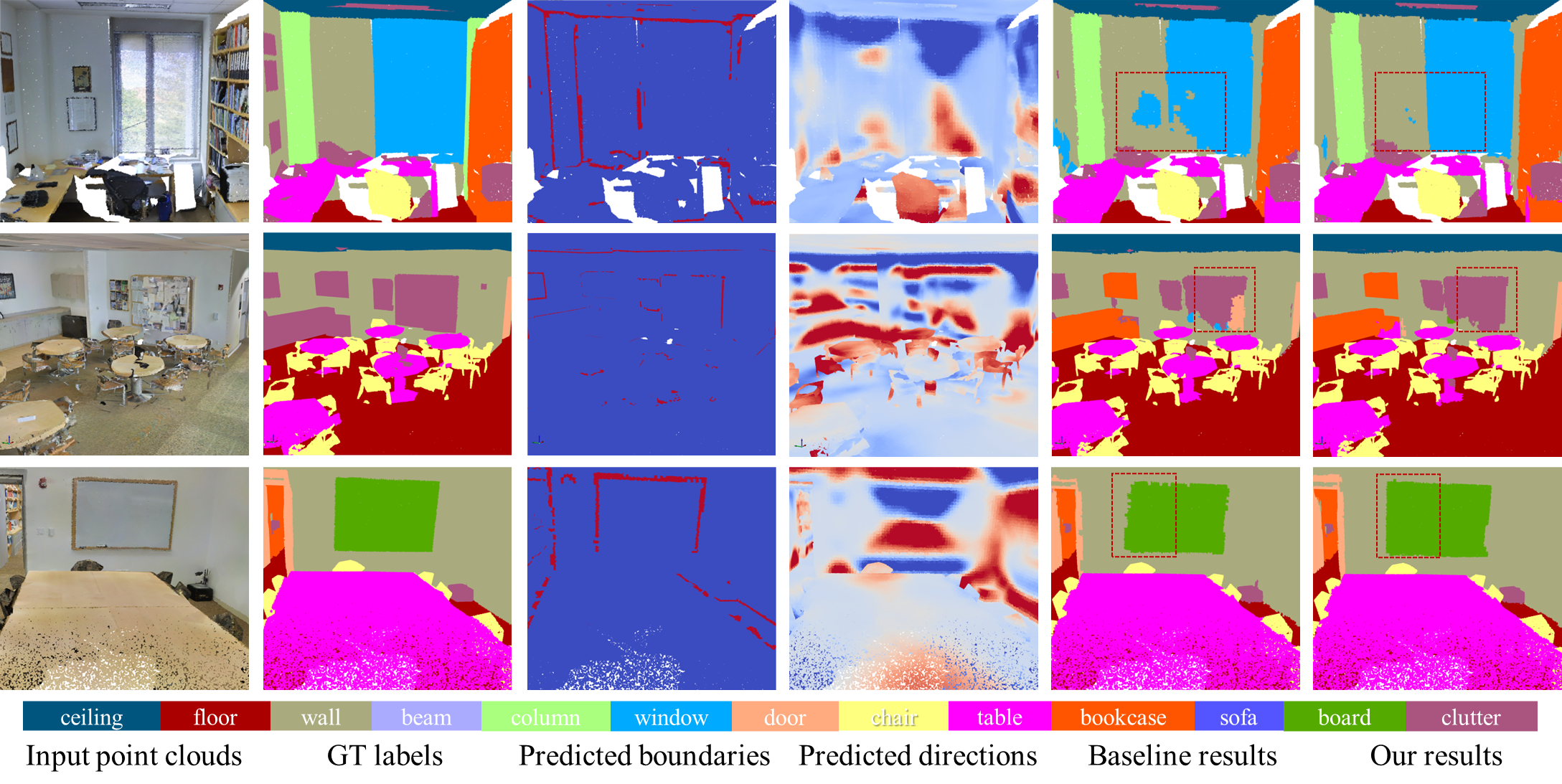}
	\end{center}
	\vspace*{-4.5mm}
	\caption{Qualitative results on the S3DIS dataset~\cite{armeni20163d}.
	KP-Conv~\cite{thomas2019kpconv} is used as our backbone, trained using 3D coordinates, colors, and normals. In the 3rd column, the predicted boundary points are rendered in red.
	In the 4th column, we visualize the Z-component (i.e., vertical direction) of the predicted vectors, where blue indicates -1 (down) and red indicates +1 (up).}
	\vspace*{-2mm}
	\label{fig:visuals3dis}
\end{figure*}

\subsection{Qualitative Evaluation}
\label{subsec:visual}
In this section, we present qualitative results of semantic scene segmentation achieved on both S3DIS~\cite{armeni20163d} and SensatUrban~\cite{hu2021towards} datasets. 

Figure~\ref{fig:visuals3dis} presents the segmentation results on the indoor dataset S3DIS.
Our joint learning and boundary-aware feature propagation strategy effectively reduce segmentation errors near boundaries for several object categories, such as window, column, and board.
To further understand why simply guiding the feature propagation with the predicted directions significantly improves the performance, we go beyond numbers by visualizing what is learned in the boundary detection and the direction prediction streams.
The visualization results reveal that even though the predicted boundaries and directions do not completely match the ground truth, they still provide informative guidance for feature propagation in the decoding layers.

\begin{figure}[t]
	\begin{center}
		\includegraphics[width=0.9\linewidth]{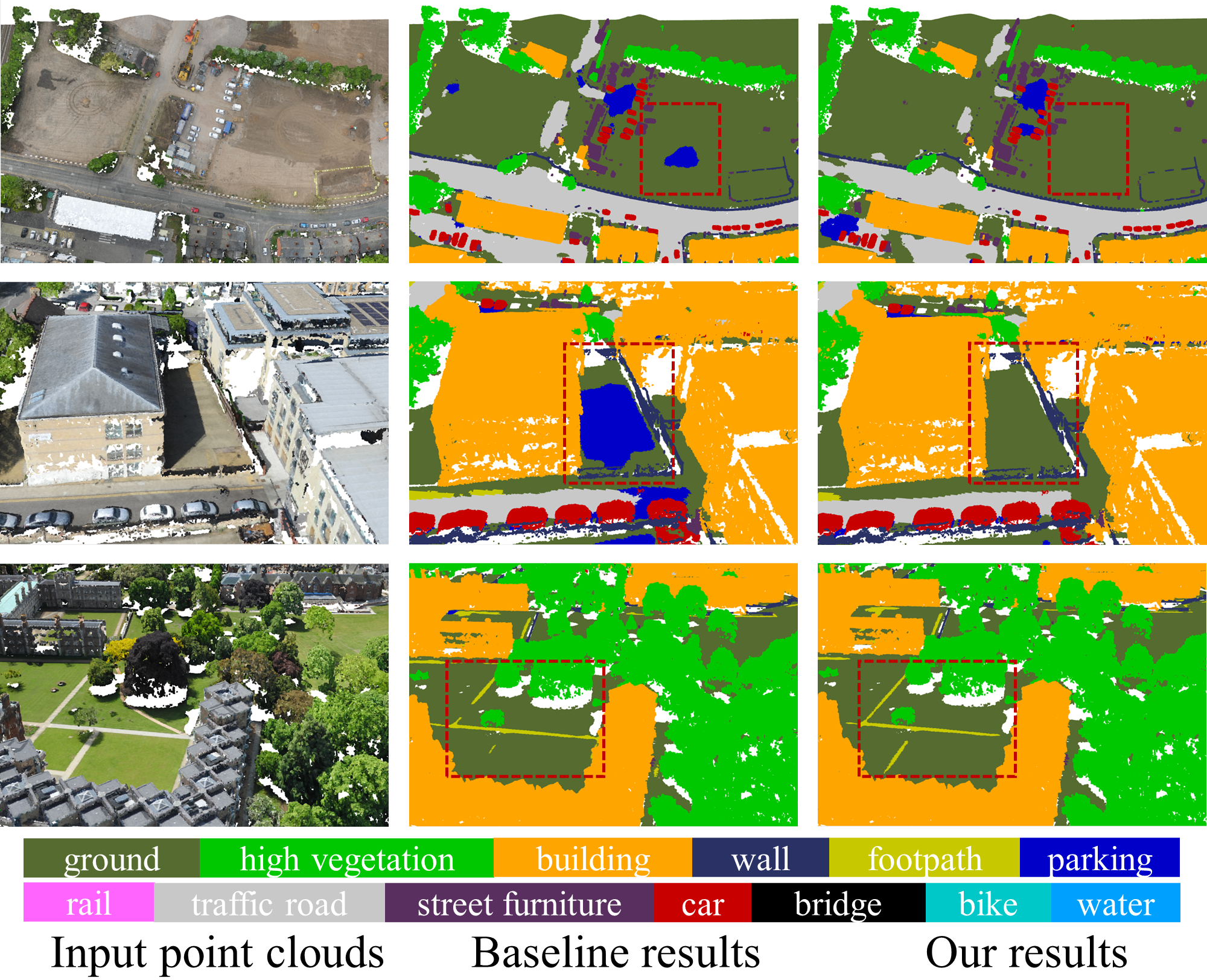}
	\end{center}
	\vspace*{-4.5mm}
	\caption{Qualitative results of semantic segmentation on the SensatUrban dataset~\cite{hu2021towards}, trained using 3D coordinates and colors.}
	\vspace*{-2mm}
	\label{fig:visualsensat}
\end{figure}

Figure~\ref{fig:visualsensat} demonstrates the segmentation results on the SensatUrban dataset, where the ground-truth labels are not presented since we do not have access to the true labels in the test set.
However, the comparison with baseline results still shows that our method achieves better boundary localization for minor categories such as parking and footpath.

\subsection{Limitations}
Our proposed approach jointly performs boundary prediction, direction prediction, and semantic segmentation.
By adopting a lightweight guiding mechanism for feature propagation, our approach can produce sharp feature maps that effectively reduce segmentation errors in boundary regions.
However, it has several limitations. First, feature propagation is performed within a neighborhood, which improves the feature consistency of object's interior on a local level.
Nevertheless, it cannot guarantee global-level feature consistency within a long range.
Second, since our work focuses on guiding the features to recover boundary information in the decoding layers, it naturally cannot cope with the information loss in the feature encoding layers.
Last, our approach requires extra pre-processing steps (i.e., generating GT boundary maps and direction maps for training).

\section{Conclusion}
We have presented a novel boundary-aware feature propagation mechanism to improve semantic segmentation of 3D point clouds, in a way such that the boundaries are pushed to the desired locations.
Our network jointly learns boundary maps, direction maps, and point-wise semantic labels end-to-end.
Extensive studies on the S3DIS and SensatUrban datasets have demonstrated the effectiveness of our approach.
Our experiments and analysis reveal two factors that contribute to the improvements in semantic segmentation.
First, the joint learning of the three tasks mutually improves the shared feature encoder.
Second, the predicted boundaries and directions are effective in guiding the points to inherit features from more homogeneous regions, which compensates for the loss of local boundary information in the decoding layers of FCNs.
Our approach is particularly effective for objects with clear geometric boundaries such as doors, windows, and street paths.

Nevertheless, the achieved improvements are still limited because the locally-performed feature propagation cannot optimize the segmentation output on a global scale.
Besides, since the three downstream tasks have different levels of complexity, combining them into a single network limits the capability of the shared feature encoder to learn discriminative features for all the tasks.
In the future, we will explore the adaptive boundary detection method~\cite{xu2021investigate} to improve semantic segmentation.
We would also like to extend our boundary-aware feature propagation mechanism using self-supervised learning techniques.

\vspace{3mm}
\noindent \textbf{\large Acknowledgement}

This work was supported by the 3D Urban Understanding Lab funded by the TU Delft AI Initiative.

{\small
\bibliographystyle{ieee_fullname}
\bibliography{egbib}
}

\end{document}